\documentclass[a4paper]{scrartcl}
\usepackage{times}
\usepackage{helvet}
\usepackage{authblk}

\usepackage[utf8]{inputenc}
\usepackage[nolist, nohyperlinks]{acronym}
\usepackage{amsmath}
\usepackage{amssymb}
\usepackage{graphicx}
\usepackage{latexsym}

\newcommand{\figref}[1]{{Fig.\,\ref{#1}}}
\newcommand{\secref}[1]{{Sec.\,\ref{#1}}}

\begin{document}

\title{Gradient-based Taxis Algorithms for Network Robotics}

\author{Christian Blum and Verena V. Hafner}
\affil{Humboldt-Universit\"at zu Berlin, Institut f\"ur Informatik\\
Kognitive Robotik, Berlin, Germany\\ email: blum@informatik.hu-berlin.de}

\date{}

\maketitle

\begin{abstract}
  Finding the physical location of a specific network node is a prototypical
  task for navigation inside a wireless network.  In this paper, we consider in
  depth the implications of wireless communication as a measurement input of
  gradient-based taxis algorithms. We discuss how gradients can be measured and
  determine the errors of this estimation. We then introduce a gradient-based
  taxis algorithm as an example of a family of gradient-based, convergent
  algorithms and discuss its convergence in the context of network robotics. We
  also conduct an exemplary experiment to show how to overcome some of the
  specific problems related to network robotics.  Finally, we show how to adapt
  this framework to more complex objectives.
\end{abstract}

\section{Introduction}

Wireless networks, ranging from cellular networks to ad-hoc networks as used in
car-to-car communication, are becoming a crucial part of our communication
infrastructure. Additionally, the paradigm shift towards the internet of things
will add wireless network capabilities to many physical objects which will then
be able to commence machine-to-machine communication \cite{Mattern2010}. These
connections can in turn span entire (ad hoc) wireless networks in between these
networked objects. One example of this already deployed today are sensor
networks.

In many scenarios for robot applications, such as using the sensing
capabilities of a decentralized sensor network or getting access to cloud based
resources, robots have to be able to integrate themselves into different
wireless networks.  In addition to these machine-to-machine communication
scenarios, mobile robots can also be nodes of very flexible, easily deployable
wireless communication networks, which can be used as communication
infrastructure.  These networks are needed for example in disaster scenarios as
a robust ad hoc communication infrastructure when local communication
infrastructure is (partly) unavailable. This approach is explored in projects
like SMAVNET \cite{Hauert2010} or AirShield \cite{Daniel2010}.

On a very basic level integration means that robot has to be in communication
range of network nodes to be able to communicate with these nodes. Because of
effects like shadowing, different spatial locations have different signal
qualities even inside the communication range. Robots have to be able to cope
with these modalities. 

Wireless communication has more degrees of freedom than wired communication,
which can be explored and exploited by a robot. Tasks like finding a position
with good reception or localizing a network node can however pose some
constraints on the network integration of the robot.

\subsection{Problem Statement}

In this paper, we consider a very basic but nevertheless very relevant task for
the robot.  Finding the physical location of a specific network node is the
prototypical task for many tasks dealing with navigation inside a wireless
network. Later in this paper we show how to adapt the presented framework to
cope with more complex objectives. 

A robot receives packets from the network node to be found at different spatial
positions and measures the signal strength of theses packets as a scalar
measurement. Thus, a scalar field is sampled at different points in space.  The
robot does not have any means to detect the direction from which the packets
arrive as would be possible with a directional antenna. Furthermore, it cannot
measure the time flight of the packets, which would enable the robot to simply
trilaterate the source. While both options are technologically feasible, we do
not want to consider them here in order to be independent of additional
technology and hardware as would be needed for such time-of-flight
measurements.

In addition to measurements of the signal strength, the robot is able to
perform local odometry measurements. They are local in the sense that they are
precise for small distances but not sufficient for global localization for
example by path integration of these measurements. Other global localization
methods like for example GPS measurements are not available to the robot.

For the sake of simplicity we assume an obstacle-free space, which is valid in
some scenarios like for example with flying robots. In other scenarios the
algorithm has to be expanded with obstacle avoidance for which there are
standard methods that are not the focus of our discussion. The algorithm
presented here converges from any starting point thus obstacle avoidance can be
implemented by stopping the taxis algorithm and restarting it after clearing
the obstacle. In general, this paper is concerned with the convergence
properties of this family of the algorithm, not with an engineering solution to
a problem. As stated, in principle the algorithm can be expanded to take
obstacles into account but we are not interested in this discussion.

\subsection{Related Work}\label{sec:RelatedWork}

A similar problem as discussed here exists in nature for example in the form of
chemotaxis \cite{Adler1966, Berg1972, Lux2004} where bacteria have to move to
the point of highest concentration of food molecules (or to the point of lowest
concentration of harmful molecules).  Because of their size, they cannot
directly measure gradients in the concentration of the molecules but they
evolved behavior suited for coping with that restriction.  These algorithms
have been adapted to the case of signal strength measurements
\cite{wadhwa2011following}. While these algorithms are effective without using
gradient information, they are less efficient because of their stochastic
nature.  Furthermore, while the algorithms proposed here are similar to the
behavior exhibited by bacteria performing chemotaxis, bacteria typically do not
have access to odometry information leading to more stochastic behavior.

The network community has developed some algorithms similar to the
gradient-based taxis algorithm discussed here. There exist for example
algorithms to calculate relative bearing from gradients \cite{Dantu2009}.
These algorithms employ, in contrast to the finite differences used here,
principal component analysis to estimate gradients. Furthermore, gradients can
be used to localize network nodes by fitting a local model to the measured
signal strength data \cite{Han2009}. This and similar algorithms need precise
position information acquired for example via GPS measurements or using laser
range finders \cite{Fink2010}. The most straightforward linear model was
explored successfully in \cite{paul2011radio, twigg2012rss}.

A taxis algorithm from the same family as the algorithm introduced here was
presented in \cite{Atanasov2012} for general abstract taxis.  This algorithm is
based on \ac{RDSA} known from the stochastic approximation literature
\cite{Kushner2003, Spall2005}. \ac{RDSA} uses a very efficient but rather
unintuitive formulation of the gradient estimation. The authors discuss the
noise characteristics of the physical model only very briefly and fail to
mention motor noise at all. As discussed later, small scale fading is
especially problematic and can violate some of the convergence conditions if
not dealt with correctly (see \secref{sec:SmallScaleFading}).  Because of that
we discuss the physical characteristics of wireless communication and the robot
in detail in the next sections.

Thus, our contribution in this paper focuses on the one hand on the gradient
estimation and its properties in the context of wireless networks.  We discuss
the implication of our use case of network robotics on the gradient estimation
and on the convergence of the taxis algorithm in detail.  In
\secref{sec:PhysModel} we discuss the nature of wireless communication with
particular attention on the proposed algorithms. We then discuss different
noise sources and their effect on the estimated gradients in
\secref{sec:GradEst}. In \secref{sec:TaxisAlgorithm} we make use of these
gradients to introduce a stochastic approximation algorithm and translate it
into the robotics world. The convergence results of this algorithm for the case
of wireless communication can then be extended to the whole family of
stochastic approximation algorithms. We then show an exemplary implementation
of the algorithm in \secref{sec:experiments} which shows how to deal with some
of the specific challenges posed by our use case.  As a last step, we adapt the
proposed algorithm to more advanced tasks in the same framework in
\secref{sec:ComplexObjectives}.

\section{Physical Model}\label{sec:PhysModel}

We discuss the underlying physical models for path loss in depth.  This forms
the basis for the discussion of the algorithm itself and is of universal
importance for all algorithms in network robotics.

\subsection{Minimalistic Physical Model}\label{sec:MinPhyModel}

The measured signal strength depends on a lot of environmental parameters and
is in general hard to calculate. In many cases, values can only be obtained
through direct measurement or numerical solution of Maxwell's Equations.

A minimalistic model of path loss, which ultimately defines the signal strength
measured by a user, is a simplified path loss model derived from the open space
path loss model. It can be written as \cite[Chap. 2]{Goldsmith2005}

\begin{equation}
  P(x) = - 10\gamma\log_{10}\left(\frac{x}{d_0}\right)\label{eq:PL}
\end{equation}

in $dB$ where $d_0$ and $\gamma$ are empirical constants and $x$ is the
distance between receiver and transmitter. $d_0$ has typical values of several
meters and $\gamma$ is about $3$ for urban environments.

Since this model is only used to get an intuition about the general behavior of
path loss and thus signal strength, we only present the one-dimensional
rotation-symmetric version of path loss. To get an intuition about its
derivatives (and thus also its gradients) the first three derivatives of the
path loss are given as

\begin{align}
  \partial_x P(x) &= -\frac{10\;\gamma}{\ln 10}\;\frac{1}{x},\\
  \partial_x^2 P(x) &= \frac{10\;\gamma}{\ln 10}\;\frac{1}{x^2},\\
  \partial_x^3 P(x) &= -\frac{20\;\gamma}{\ln 10}\;\frac{1}{x^3}. \label{eq:ThirdDiffPL}
\end{align}

This model can be motivated as physically plausible for the case of free space
with some effective signal attenuation.

\subsection{Elaborate Physical Model}\label{sec:ElaboratePhysModel}

In general, Maxwell's equations have to be solved to correctly calculate radio
wave propagation. With complete knowledge of the environment this is indeed
possible but practically almost never feasible because of the high
computational costs and because complete knowledge is unobtainable in real
scenarios.

As a consequence, effective radio propagation models have been developed.  In
general, there are three components which constitute these models. Not all
models make use of all components depending on the purpose of the model.  These
three components are the following \cite{Goldsmith2005}:

\begin{itemize}
  \item path loss (large scale)
  \item shadowing (medium scale)
  \item fading (small scale)
\end{itemize}

In the following paragraphs these components are discussed with a robot as a
network node in mind.

Large scale path loss is the simplified path loss model discussed in
\secref{sec:MinPhyModel}.  The most straightforward case for this is the free
space model which takes only the most basic physical effect, i.e., wave
propagation in free space, into account. More elaborate models usually
approximate all kinds of empirical effects on path loss by adjusting the
exponent of the path loss function \eqref{eq:PL}. This exponent is tuned to
different scenarios via empirical measurements. For the following discussion on
taxis algorithms, it is important to note that these models are strictly
monotonically decreasing.

Shadowing is the effect of large obstructions such as a hill or wall in the
direct path of wave propagation. In some cases, such as the standard double
plasterboard wall in an office environment, these effects can simply be
measured and added as additional path loss. More complex scenarios need more
complicated models. Note that shadowing can only attenuate signals. 

Some configurations of obstructions, consisting for example of walls of
different attenuation, can create situations in which a robot simply following
the proposed algorithms can get stuck because the signal strength gradient
would try to guide the robot through a wall. An obstacle avoidance algorithm,
which needs some degree of knowledge about the environment, has to be used to
help the robot escape these situations. As stated earlier, the taxis algorithm
itself converges regardless of starting point so it can be stopped when the
robot starts the obstacle-avoidance algorithm and restarted once it has cleared
the obstacle.

Fading is a result of multipath propagation of radio waves. Superposition of
waves traveling different paths interfere because of their different phases on
a physical level either constructively or destructively. This leads to
spatially varying signal strengths on the length scale of the wave length of
the radio signal (about $12.5\; cm$ for $2.4\;GHz$). Fading will be discussed
in detail in \secref{sec:SmallScaleFading}.

Taking these three effects together, a robot measuring the signal strength of a
fixed receiver deals with a strictly monotonically increasing function with a
considerable amount of noise and spatially varying characteristics. The signal
strength is in general --- because of shadowing --- a non-monotonic function of
distance to the transmitter, preventing direct distance estimations using only
signal strength measurements.

For moving robots, Doppler shift \cite{Goldsmith2005} can be an additional
factor.  In general, Doppler shift does not affect signal strength on a global
level. However, for a sender-receiver pair working in some specified frequency
band, Doppler shift can lead to a reduced received signal strength by shifting
part of the signal to frequencies which are outside of the communication
channel bandwidth.  Doppler shift is dependent among others on the speed of the
robot, carrier frequency and modulation. Because of this it is difficult to
give a general answer to how important this effect is. However, most robots are
capable of holding their position --- for example hovering flying robots ---
for the time it takes to make a measurement. This works trivially for ground
based robots but also for some flying robots but is an issue for flying
robots based on the fixed wing principle.

\subsection{Small Scale Fading}\label{sec:SmallScaleFading}

\begin{figure}[t]
  \centering
  \includegraphics[width=0.8\textwidth]{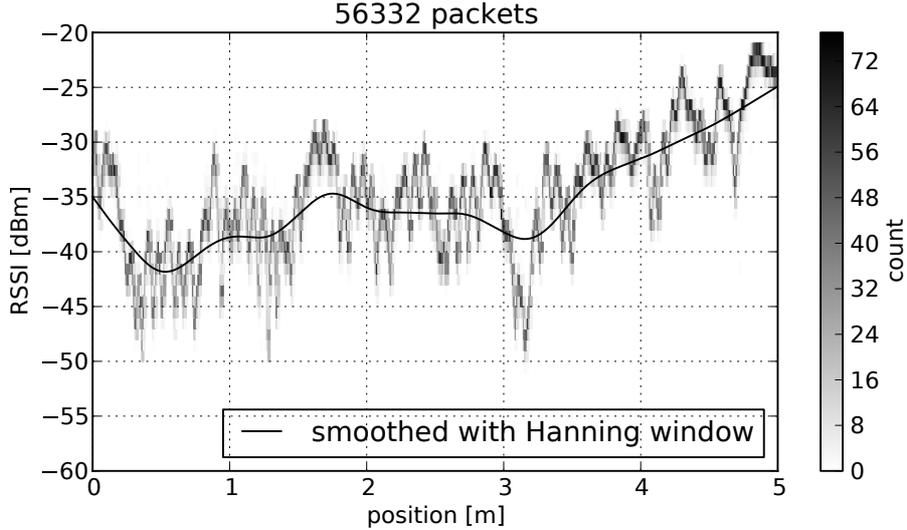}
  \caption{Example indoor measurement of small scale fading caused by walls and
    other scatterers. A robot was moving along an arbitrary line through the
    scalar signal strength field of a network node. The distribution of the
    signal strength of the measured packages as well as smoothed version of the
    data is depicted.}
  \label{fig:smallScaleFading}
\end{figure}

As discussed in the previous section, small scale fading is the result of
interference and has an effect on the length scale of the wavelength. It leads
to fluctuations with periodic character with a periodicity of the wavelength of
the wireless signal (which is dependent on the used channel, i.e., the base
frequency). This effect is deterministic and can lead to local minima in the
signal strength, which would violate some convergence conditions (see
\secref{sec:FDSAConvergence}). 

An example measurement is depicted in \figref{fig:smallScaleFading}. The
deterministic fluctuations on the scale of the wavelength ($12.5\; cm$ for
$2.4\;GHz$ in this case) can clearly be seen.  These effects can be stronger
than the noise sources and have to be dealt with in order to ensure convergence
of the presented algorithms.

For a flying robot, this is typically not a problem because it cannot hold
position with a precision of this magnitude. This means that because of the
stochastic movement (induced by aerodynamics and external factors like wind)
these deterministic fluctuations are turned into additional noise. For ground
based robots, this effect has to be mitigated by for example taking several
samples and averaging over an area larger than the wavelength.  This will also
be demonstrated in our experiments in \secref{sec:experiments}.

\subsection{Abstract Model: Notation}\label{sec:AbstractModel}

Based on the more elaborate models discussed above, we describe an abstract
model to be used for a compact notation in the remaining part of the paper. It
encompasses all previously discussed more complex models and is rather vague to
enable the short notation. It should be noted that it implicitly contains all
characteristics needed for the convergence discussion.

In our model, we can measure the true signal strength $f(\vec{x})$ only up to a
measurement error

\begin{equation}
  f_{\text{mes}}(\vec{x}) = f(\vec{x}) + \epsilon(\vec{x}) \label{eq:MesModel}
\end{equation}

where the measurement error $\epsilon(\vec{x})$ is a random variable with

\begin{align}
  E[\epsilon(\vec{x})] &= 0,\\
  V[\epsilon(\vec{x})] &= \sigma^2.
\end{align}

The measurement errors $\epsilon(\vec{x})$ are i.i.d. for the individual
measurements.  Two things should be noted: Firstly this definition does not
make any assumptions about the distribution as long as the two conditions above
are met and secondly that this condition can be violated by small scale fading
(see \secref{sec:SmallScaleFading}) if not dealt with correctly.  The signal
strength function $f(\vec{x})$ is a $p$-dimensional scalar field with one
global maximum.

This model is agnostic in regard to the actual characteristics of the function
$f(\vec{x})$.  When discussing the convergence properties of the proposed
algorithm in \secref{sec:FDSAConvergence}, we discuss its characteristics. For
now it serves as a pure notational convenience.

\section{Gradient Estimation}\label{sec:GradEst}

\subsection{Central Differences}

We estimate the true gradient

\begin{equation}
  \vec{g}(\vec{x}) = \nabla f(\vec{x})
\end{equation}

with central differences ($i$th component):

\begin{equation}
  \hat{\vec{g}}_i(\vec{x}) = 
  \frac{f(\vec{x} + h\vec{e}_i) - f(\vec{x} - h\vec{e}_i)}{2h}, 
  \label{eq:GradEstimation}
\end{equation}

where $\hat{\vec{g}}(\vec{x})$ denotes the estimate of the true gradient
$\vec{g}(\vec{x})$, $\vec{e}_i$ is the $i$th unit vector and $h$ is the used
stepwidth. This estimation is the most standard one with improved precision in
relation to one-sided finite differences.  Right now this estimation does not
take into account the measurement errors, which will be discussed in
\secref{sec:GradientErrorAnalysis}, but only the errors from the numerical
approximation.

Using second order Taylor expansions \cite{Bronstein2008}

\begin{equation}
\begin{split}
  f(\vec{x} \pm h\vec{e}_i) &= f(\vec{x}) + \epsilon(\vec{x}) \\
  &\pm h \partial_{x_i} f(\vec{x}) \\
  &+ \frac{h^2}{2} \partial_{x_i}^2 f(\vec{x}) \\
  &\pm \frac{h^3}{6} \partial_{x_i}^3 f(\vec{x} \pm \xi h \vec{e}_i)
\end{split}
\end{equation}

with $0 < \xi < 1$ yields for the gradient estimate

\begin{equation}
  \hat{\vec{g}}_i(\vec{x}) = \partial_{x_i} f(\vec{x}) + 
  \frac{h^2}{12} 
  \left( 
    \partial_{x_i}^3 f(\vec{x}  + \xi_1 h \vec{e}_i) 
    - \partial_{x_i}^3 f(\vec{x} - \xi_2 h \vec{e}_i)
  \right)
\end{equation}

where $0 < \xi_1 < 1$, $0 < \xi_2 < 1$.

\subsection{Error Analysis}\label{sec:GradientErrorAnalysis}

A gradient estimate calculated with measured data contains, additionally to the
numerical errors discussed above, measurement errors. We are using the notation
$f_{\text{mes}}(\vec{x}) = f(\vec{x}) + \epsilon(\vec{x})$ as described in the
abstract model in \secref{sec:AbstractModel} to write the estimated gradient
based on measurements $\hat{\vec{g}}_{\text{mes}}(\vec{x})$ as

\begin{equation}
\begin{split}
  \hat{\vec{g}}_{\text{mes},i}(\vec{x})
    &= \frac{
      f_{\text{mes}}(\vec{x} + h\vec{e}_i) 
      - f_{\text{mes}}(\vec{x} - h\vec{e}_i)
    }{2h}\\
    &= \hat{\vec{g}}_i(\vec{x}) 
      + \frac{
        \epsilon(\vec{x} + h\vec{e}_i) 
        - \epsilon(\vec{x} - h\vec{e}_i)
      }{2h}.
\end{split}
\end{equation}

The expectation value and variance of $\hat{\vec{g}}_{\text{mes}}$ can now be calculated as

\begin{align}
  E&[\hat{\vec{g}}_{\text{mes},i}(\vec{x})] 
    = \partial_{x_i} f(\vec{x}) + \frac{h^2}{12} 
      \left( 
        \partial_{x_i}^3 f(\vec{x} + \xi_1 h \vec{e}_i) 
        - \partial_{x_i}^3 f(\vec{x} - \xi_2 h \vec{e}_i) 
      \right)\\
  V&[\hat{\vec{g}}_{\text{mes},i}(\vec{x})] 
    = \frac{\sigma^2}{2h^2}.\label{eq:FDVar}
\end{align}

As expected, the relation $E[\hat{\vec{g}}_{\text{mes}}(\vec{x})] =
\vec{g}(\vec{x}) + O(h^2)$ holds for the expectation value of the estimated
gradient. Estimating the gradient using central differences without measurement
errors yields the same relation for the expectation value.

There are two kinds of errors contained in
$\hat{\vec{g}}_{\text{mes}}(\vec{x})$, namely a numerical error produced using
finite differences which behaves like $O(h^2)$ and a stochastic error due to
the measurement error $\epsilon(\vec{x})$ which behaves like $O(h^{-2})$.

\subsection{Motor Noise: Measurement Errors}\label{sec:MotorNoiseFD}

For a real robot, there is, additionally to the physical measurement noise
$\epsilon$ as in \eqref{eq:MesModel}, so called motor noise. This noise is
added to any motor command and thus affects all movements. In general this
noise is vectorial and affects every infinitesimally small movement of the
robot thus leading to a random walk-like behavior. This behavior is depicted in
the inset of \figref{fig:Iteration} in a graphical way.

Since it does not make a qualitative difference for our argument \footnote{The
argument is purely statistical and no fixed coordinate system is used, so the
actual end position of the robot is not important.} but simplifies notation, we
abstract here from this vectorial noise to a noise in the direction of movement
only.  For the central differences this means that $f(\vec{x})$ is not sampled
at $f(\vec{x}+h\vec{e}_i))$ but at $f(\vec{x}+(h+\epsilon_h)\vec{e}_i))$. This
yields a gradient estimation of

\begin{equation}
  \hat{\vec{g}}_{\text{mes},i}(\vec{x}) 
    = \frac{1}{2h} 
      \left( 
        f_{\text{mes}}(\vec{x} + (h+\epsilon_{h,1})\vec{e}_i)
        - f_{\text{mes}}(\vec{x} - (h+\epsilon_{h,2})\vec{e}_i)
      \right)
\end{equation}

The $\epsilon_h$ are assumed to be i.i.d. and bias-free\footnote{The specific
distribution has no influence on the results of the discussion here as long as
it is normalizable. Even a bias in the distribution, as for example the robot
always moving a bit more to the left of its movement axis, has no influence
since in the real algorithm, the movement direction is a stochastic quantity
which changes every iteration step.}.  In contrast to a real random walk, the
end positions are only distributed with a one-dimensional distribution in the
direction of movement. Writing the full vectorial distributions would clutter
the formulas but not change our results so we only discuss the simplified case.
For small $\epsilon_h$ we can expand the $f(\vec{x}+(h+\epsilon_h)\vec{e}_i)$
with

\begin{equation}
  f(\vec{x}+(h+\epsilon_h)\vec{e}_i) = 
    f(\vec{x}+h\vec{e_i})
    + \epsilon_h\partial_{x_i} f(\vec{x}+h\vec{e}_i) 
    + O(\epsilon_h^2).
\end{equation}

For small $\epsilon_h$ this expansion is valid and effectively yields a larger
measurement error of $f(\vec{x})$ than the pure physical measurement error
$\epsilon$. For further analysis we can absorb this additional error in the
measurement error as long as their distributions are similar. In general, the
central limit theorem \cite[p. 129]{Blobel1998} holds and allows us to simply
add both errors \footnote{While the path loss noise is often modeled by
Rayleigh or Rician fading models, which result in non-Gaussian noise, these
models are stochastic approximations to multipath fading. Multipath fading
itself however is deterministic and only the measurement errors themselves are
stochastic and Gaussian (see also \secref{sec:SmallScaleFading}).}

\subsection{Signal-To-Noise Ratio}

The \ac{SNR} is a useful quantity describing the relation of a measured signal
amplitude to the amplitude of the noise introduced into this channel. In the
case of gradient estimation it can be used to characterize the quality of the
estimated gradients in terms of a (virtual) sensor reading.

Defining the \ac{SNR} of some function $e(\vec{x})$ as \cite[Chap. 2]{Smith1997}

\begin{equation}
  S\!N\!R = \frac{E[e(\vec{x})]}{\sqrt{V[e(\vec{x})]}}
\end{equation}

and applying this definition for the $i$th component of the estimated gradient 
$\hat{\vec{g}}_{\text{mes},i}(\vec{x})$ yields

\begin{equation}
  S\!N\!R_i =  \frac{\sqrt{2}h}{\sigma}\partial_{x_i} f(\vec{x})
  +\frac{\sqrt{2}h^3}{12\sigma}
  \left( 
    \partial_{x_i}^3 f(\vec{x} + \xi_1 h \vec{e}_i) 
    - \partial_{x_i}^3 f(\vec{x} - \xi_2 h \vec{e}_i) 
  \right)
\end{equation}

which can be approximated for small $h$ with

\begin{equation}
  S\!N\!R_i\sim \frac{h}{\sigma}\; \partial_{x_i} f(\vec{x}).
\end{equation}

For large $h$ the error due to the remainder term of the Taylor expansion,
which is the bias of the gradient estimation, dominates. The \ac{SNR} as
defined above then loses its meaning.

For the specific problem of path loss \eqref{eq:PL} as the function for which
the gradient is estimated, the third derivative of the path loss
\eqref{eq:ThirdDiffPL} vanishes like $\frac{1}{x^3}$.  This means that far from
the source this approximation is valid also for large $h$ since the bias only
increases like $h^2$.  \section{Taxis Algorithm}\label{sec:TaxisAlgorithm}

The robot estimates gradients using finite differences by sampling the scalar
field and uses this gradient as the direction for the next step. We formally
introduce this algorithm and discuss its convergence and statistical
properties.


\subsection{Finite Difference Stochastic Approximation}\label{sec:FDSA}

\begin{figure}[t]
  \centering
  \includegraphics[width=0.6\textwidth]{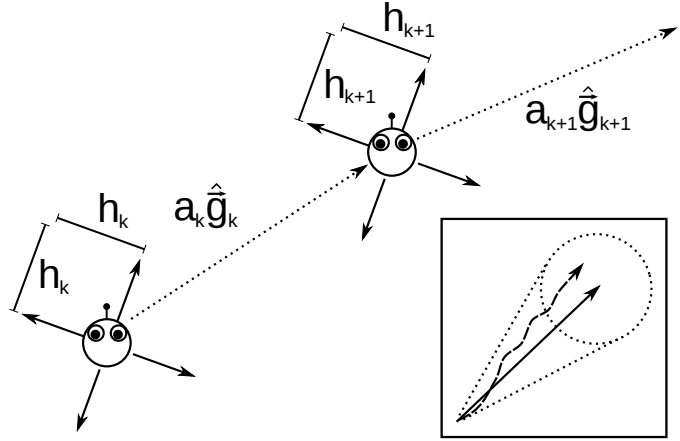}
  \caption{The sequence of two iterations. The cross of arrows denoted with
    lengths of $h_k$ show the steps taken for gradient estimation. The dotted
    arrows denoted with $a_k$ show the actual iteration steps. The inset shows
    the effect of motor noise. The step to be taken is depicted as a solid
    arrow, the approximate distribution of end position with a dotted circle
    and one of the realizations of movement with motor noise as a dashed
    arrow.}
  \label{fig:Iteration}
\end{figure}

One of the most basic algorithms to find the minimum of some scalar field is
the method of steepest descent. It consists of calculating (or in this case
estimating) the gradient $\hat{\vec{g}}_{\text{mes}}(\vec{x})$ beginning at
some starting point $\hat{\vec{x}}_0$ and following it with some stepwidth
$a_k$:

\begin{equation}
  \hat{\vec{x}}_{k+1} = 
  \hat{\vec{x}}_k + a_k\hat{\vec{g}}_{\text{mes}}(\hat{\vec{x}}_k).
\end{equation}

The stepwidths $h$ of the gradient estimation \eqref{eq:GradEstimation} may
also be adapted for each step and are thus denoted as $h_k$. The iterates of
this algorithm are estimates of the position of the minimum, starting from an
initial guess $\hat{\vec{x}}_0$, which is in this case trivially the position
of the robot when the algorithm is started, and are therefore denoted as
$\hat{\vec{x}}_k$.  Thus, this algorithm is basically stateless and the current
position of the robot is always the best estimate of the minimum. The gradients
then always point the way in which the robot has to move to reach the minimum.
Two iteration steps of this algorithm are depicted in \figref{fig:Iteration} in
a graphical way.

This algorithm has been subject to theoretical and numerical analysis since the
1950s because it can be used to solve common optimization tasks in many
scientific and engineering areas. This particular method based on central
differences is known as \ac{FDSA} in the field of Stochastic Approximation
\cite{Spall2005}.

\subsection{Convergence}\label{sec:FDSAConvergence}

It can be shown that this algorithm converges if $a_k$, $h_k$, $f(\vec{x})$ and
$\epsilon(\vec{x})$ conform to some conditions \cite[pp. 159-162]{Spall2005}.
Informally speaking these conditions demand that

\begin{itemize}
  \item $f(\vec{x})$ has a global minimum at $\vec{x}^\star$,
  \item $\epsilon(\vec{x})$ has a mean of zero and finite variance,
  \item the Hessian Matrix 
  $H(\vec{x}) = \frac{\partial^2 f(\vec{x})}{\partial \vec{x} \partial \vec{x}^\top}$ 
  exists and is uniformly bounded for all $\vec{x}$.
\end{itemize}

The first condition is discussed in \secref{sec:ElaboratePhysModel}  and is
found to be satisfied for the case of signal strength as the scalar field from
which the samples used to estimate the gradient are taken.  It has one global
maximum which can easily be turned into a global minimum by multiplying the
measured values by minus one.

The second condition constrains the noise of the signal. Measurement noise as
modeled in \secref{sec:AbstractModel} has mainly physical origins and is
discussed in \secref{sec:ElaboratePhysModel}.  Additionally, there is noise
originating from the motors of the robot as discussed in
\secref{sec:MotorNoiseFD}.  Finite variance of the noise is easily satisfied by
all real systems.  Small scale fading can be a deterministic bias for the
signal strength measurements as discussed in \secref{sec:SmallScaleFading}, but
that bias can be dealt with as shown in \secref{sec:experiments}.  The bias of
motor noise has been discussed to be zero in \secref{sec:MotorNoiseFD} because
of the stochastic nature of the iterate.

The third condition --- in simplified terms --- requires the scalar field to be
smooth.  This constraint is satisfied because path loss is a physical effect
and physical fields governed by Maxwell's equations always fulfill this
smoothness condition.

Furthermore (and with the most practical relevance), gain sequences $a_k$ as
well as the sequences of step sizes $h_k$ for the central differences are
restricted for a convergent algorithm:

\begin{align}
  a_k &> 0,\\
  h_k &> 0,\\
  a_k &\rightarrow 0,\\
  h_k &\rightarrow 0,\\
  \sum_{k=0}^\infty a_k &= \infty,\\
  \sum_{k=0}^\infty a_k h_k &< \infty,\\
  \sum_{k=0}^\infty \frac{a_k^2}{h_k^2} &< \infty.
\end{align}

Thus, $h_k \rightarrow 0$ slower than $a_k$. These conditions have to be
satisfied by a practical implementation of this algorithm. However, these
cannot be considered as design guidelines since they only restrict the design
space of the algorithm.  The next section can give more insight into the choice
of parameter sequences for $a_k$ and $h_k$.

\subsection{Distribution of the Iterate}

Unfortunately, there is no known finite-sample ($k<\infty$) distribution for
$\hat{\vec{x}}_k$ for general nonlinear problems \cite[p. 112]{Spall2005}. But
asymptotic ($k\rightarrow\infty$) normality of this distribution can be shown
for more specific choices of $a_k$ and $h_k$ \cite[p. 162-164]{Spall2005}:

\begin{align}
  a_k &= \frac{a}{(k+1+A)^\alpha},\\
  h_k &= \frac{h}{(k+1)^\gamma}.
\end{align}

Here $a>0$, $h>0$, $\alpha>0$, $\gamma>0$ is assumed. $A\geq 0$ is a stability
constant which ensures small enough gains in the beginning and large enough
gains in the end.

In order to show asymptotic normality, some constraints on these constants have
to be added. The most practically relevant ones are:

\begin{align}
  \beta~\equiv \alpha - 2\gamma &> 0,\\
  3\gamma - \frac{\alpha}{2} &\geq 0.
\end{align}

If these conditions are satisfied, asymptotic normality of $\hat{\vec{x}}_k$
can be shown. The forms of the mean and variance of the resulting Gaussian
distribution are unwieldy but closed-form expressions of both can be found in
literature.

The rate of stochastic convergence of $\hat{\vec{x}}_k$ to $\vec{x}^\star$ is
then proportional to $k^{-\frac{\beta}{2}}$. $\beta$ is maximized at $\alpha=1$
and $\gamma=\frac{1}{6}$ leading to a maximal attainable stochastic convergence
rate of $k^{-\frac{1}{3}}$. This can in turn serve as a general guideline for
the choice of the parameter sequences. Further details and design guides for
the practical choice of the series can be found in \cite{Spall2005}.

In principle the particular choice of the series $a_k$ and $h_k$ weighs
exploration against exploitation. These parameter series do not have to be
analytic, though. Consequently, they can be adaptive as long as the conditions
discussed in \secref{sec:FDSAConvergence} are met.  With this step ideas from
for example infotaxis \cite{Moraud2010} can be incorporated.

\subsection{Motor Noise: Iteration Steps}

In addition to the effect of motor noise discussed in
\secref{sec:MotorNoiseFD}, the error in the actual movement of the robot when
iterating the algorithm has to be considered.

Motor noise in the movement of the robot --- as long as it is truly random and
not biased --- can be thought of as an additional error in the estimated (or in
a way measured) gradient of $f(\vec{x})$. This discussion is similar to the one
in \secref{sec:MotorNoiseFD}.

This error is constant per unit length but because the stepwidth $a_k$ is
decreasing, the resulting total movement error is also decreasing. Since the
error due to the central difference approximation \eqref{eq:FDVar} increases
with decreasing $h_k$, the movement error becomes insignificant for large $k$,
thus not influencing convergence.

\subsection{Other Stochastic Approximation Algorithms}

\ac{FDSA} is only one algorithm of a family of stochastic approximation
algorithms which have similar characteristics and similar convergence
conditions. It was chosen because of the intuitive formulation of the gradient
estimation which enabled us to discuss the properties of the estimated
gradients in the light of our use case of network robotics.

Other algorithms from this family work with different formulations for the
gradient estimation and can be proved analytically to converge. They are known
from the stochastic approximation literature \cite{Spall2005} and have been
used successfully in robotics contexts \cite{Atanasov2012}.

%

The detailed conditions for convergence of these algorithms differs but the
basic constraints for the measured scalar field as stated in
\secref{sec:FDSAConvergence} are the same for all algorithms from this family.
Thus, the discussion related to wireless communication in
\secref{sec:FDSAConvergence} is valid for the whole family of algorithms.

Then there are a lot of ad hoc gradient based algorithms, for which the
discussion on the noise and the estimated errors in \secref{sec:PhysModel} and
\secref{sec:GradEst} respectively can be applied or are very similar, but for
which no formal proofs of convergence exist.  The discussion here can however
be a good guideline for the design of such algorithms.

\section{Experiments}\label{sec:experiments}

\begin{figure}[h]
  \centering
  \includegraphics[width=0.6\textwidth]{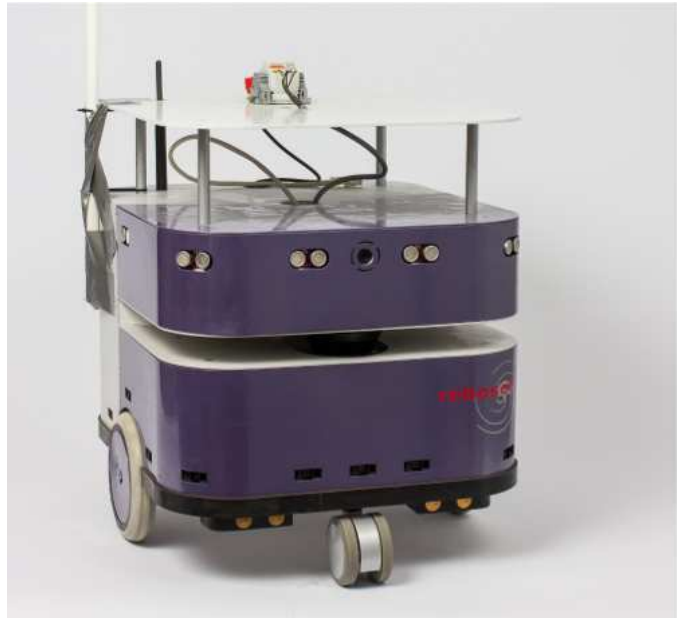}
  \caption{The robot, a Robosoft RobuLAB-10, used in the experiments}
  \label{fig:robulab}
\end{figure}

In order to show how to overcome the problems of small scale fading we
implemented the algorithm with a ground based robot for which the effects of
small scale fading are much more severe than for a flying robot since it can
position itself with a much higher precision (also see the discussion in
\secref{sec:SmallScaleFading}).

The algorithm was first implemented with four measurements for every gradient
estimation following its original formulation. The resulting behavior was
random-walk like and did not show any convergence. The most likely reason for
this behavior is small scale fading because of its local periodic nature and
the resulting effect on this naive gradient estimation.

Since a robot has to move from one sampling point to the next while executing
the algorithm, we decided to exploit this movement and measure continuously
while moving along the two axes of the gradient estimation. We supplemented the
finite difference estimation of the derivatives with a linear model of the
data collected along these lines, which basically uses the same local
linearity assumption as finite differences. The model is fitted to the data
collected while moving and used as an estimate of the derivative.  If the
model is fitted over an interval larger than the wavelength of the wireless
communication, the impact of small scale fading is mitigated.

We used a Robosoft RobuLAB-10 carrying a laptop with an attached wifi card as
our robot and a second laptop with the same configuration as the network node.
We used 802.11g as the communication standard and all antennas were positioned
well above the ground and above the robot using plastic rods. The robot is
depicted in \figref{fig:robulab}. 

\begin{figure}[h]
  \centering
  \includegraphics[width=0.7\textwidth]{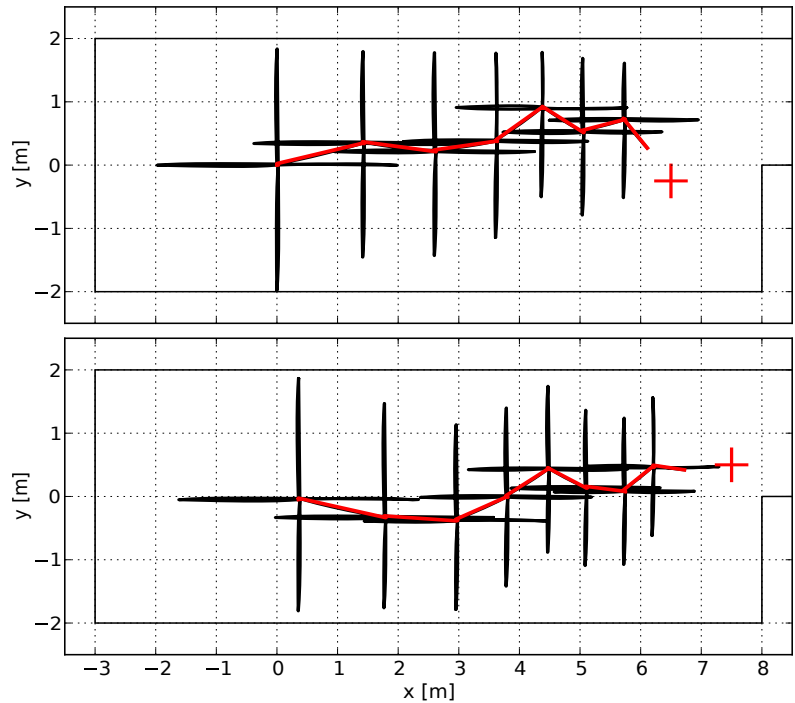}
  \caption{Depicted are the complete trajectories of the robot in black, the
    position of the measured network node as a red cross and the physical
    boundaries of the room as a thin black line for two experiments.  The
    trajectories of the estimates of the maximum are highlighted in red.}
  \label{fig:exampleRun}
\end{figure}

The trajectories of two indoor example runs of the experiment are depicted in
\figref{fig:exampleRun} and show good convergence of the position of the robot
towards the network node. We chose an indoor scenario because the effect of
small scale fading is enhanced by a lot of scatterers like walls, furniture or
metal doors

This experiment shows that the effects of small scale fading can be dealt with
very well even in this worst case of a ground based robot in an indoor scenario
with lots of multipath effects.

\section{Adapting more Complex Objectives}\label{sec:ComplexObjectives}

\begin{figure}[h]
  \centering
  \includegraphics[width=0.7\textwidth]{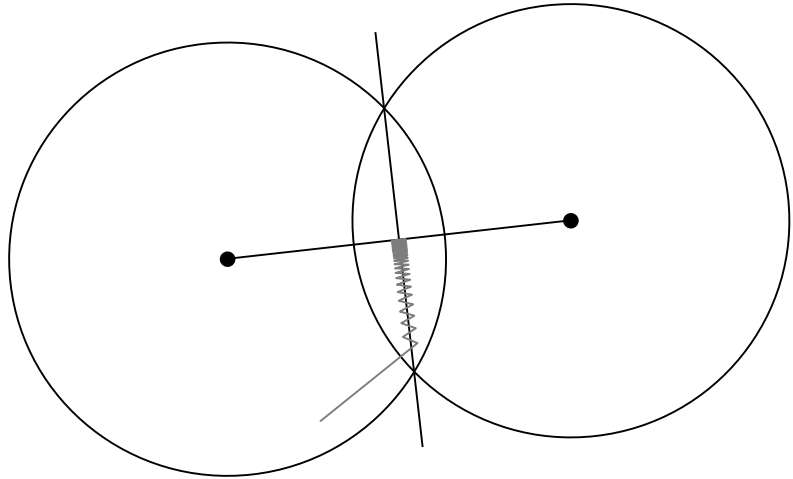}
  \caption{Two network nodes are depicted as black circles. The two lines
    represent the two parts of the objective function, the points of maximal
    sum of both signal strengths and the points of equal signal strength,
    respectively.  The hypothetical example run is depicted as a grey line.}
  \label{fig:Complex}
\end{figure}

More complex objectives than finding a network node can be addressed using the
proposed algorithm. For this, a target function, which satisfies the conditions
stated in \secref{sec:FDSAConvergence} has to be constructed in terms of signal
strength measurements only.

As an example objective we choose a typical task in network robotics, namely
bridging two network nodes, i.e., moving to the position between the nodes with
maximum signal strength and equal signal strength to both nodes. This task
often occurs when a gap in the network has to be bridged.

For this task we define the objective function $g(\vec{x})$ as

\begin{equation}\label{eq:ComplexObjectiveFunction}
  g(\vec{x}) = \| f_1(\vec{x}) - f_2(\vec{x}) \| -  \| f_1(\vec{x}) + f_2(\vec{x}) \|  \\
\end{equation}

which can be written (assuming $f(\vec{x}) >= 0$ which is true for signal
strengths) as

\begin{equation}
  g(\vec{x})=\begin{cases}
    -2 f_2(\vec{x}) &f_1(\vec{x}) > f_2(\vec{x})\\
    -2 f_1(\vec{x}) &f_2(\vec{x}) > f_1(\vec{x})\\
    -\|f_1(\vec{x}) + f_2(\vec{x}) \| &f_1(\vec{x}) = f_2(\vec{x})
  \end{cases}
\end{equation}

and only consists of signal strength measurements. This objective function has
one global minimum at the desired point with  properties as stated above.  A
hypothetical example run of this algorithm is depicted in \figref{fig:Complex}.

In principle, the behavior of this algorithm shows two stages. First the robot
moves towards the line --- in two dimensions, in three dimensions this is a
plane --- of equal signal strength to both nodes and then oscillates about this
line towards the point of maximum signal strength.  This can be seen as a
general gradient descent in the first stage and then a line search --- or a one
dimensional gradient descent --- in the reduced space of this line.

This objective function has the same properties as the signal strength itself,
satisfying the constraints of the algorithm. This is true because noise is
added before the absolute value is calculated. However, close to the area equal
signal strength of both nodes, both signal strengths cancel out in the first
term of \eqref{eq:ComplexObjectiveFunction}. This results in the absolute value
to be only calculated from the noise, thus positively biasing it which in turn
violates the constraint on the noise of having an expectation value of zero.
Thus, in this area convergence cannot be guaranteed analytically. Nevertheless,
this is not of any practical relevance because the problem reduces to a
lower-dimensional gradient descent algorithm in this area.

\section{Conclusion}

This paper showed that network robotics can make use of algorithms based on
stochastic approximation, working with signal strength measurements, for tasks
like navigation. We established estimates of the precision of gradients
calculated from signal strength measurements.  Signal strength measurements as
well motor noise was physically motivated and its effects on the convergence of
these algorithms was discussed in depth. The algorithm was also implemented
experimentally to show how to deal with some of the specific challenges posed
by network robotics.  Additionally, we showed that more complex objectives can
be formulated in this framework

\section*{Acknowledgements} We thank everybody who contributed to the success
of this project. This includes all members of the Cognitive Robotics group and
the DFG graduate research training group METRIK (GRK 1324), which also funds
one of the authors. 

\bibliographystyle{unsrt}
\bibliography{paper}

\begin{acronym}
  \acro{FDSA}{Finite Difference Stochastic Approximation}
  \acro{SPSA}{Simultaneous Perturbation Stochastic Approximation}
  \acro{RDSA}{Random Direction Stochastic Approximation}
  \acro{SNR}{signal-to-noise ratio}
\end{acronym}

\end{document}